\documentclass[runningheads]{llncs}
\usepackage{graphicx}
\usepackage{comment}
\usepackage[acronym]{glossaries}
\usepackage{algpseudocode}
\usepackage{algorithm2e}
\usepackage{tabularx}
\usepackage{adjustbox}
\usepackage{mathtools}
\usepackage{amsmath,amssymb}
\usepackage{resizegather}
\usepackage{booktabs}
\usepackage{multirow}
\usepackage{graphicx}
\usepackage{subfig}
\usepackage{hyperref}               
\hypersetup{
    colorlinks=true,
    linkcolor=blue,
    filecolor=magenta,      
    urlcolor=magenta,
    citecolor=blue,
    pdftitle={Overleaf Example},
    pdfpagemode=FullScreen,
    }
\usepackage{xcolor}

\newcommand\tocite[1]{\textcolor{blue}{[REFERENCE]}}

\newacronym{ddpm}{DDPM}{Denoising Diffusion Probabilistic Models}
\newacronym{gan}{GAN}{Generative Adversarial Networks}

\begin{document}
\title{WordStylist: Styled Verbatim Handwritten Text Generation with Latent Diffusion Models}

%
\titlerunning{WordStylist: Styled Verbatim Handwritten Text Generation with LDM}
%
\author{Konstantina Nikolaidou\inst{1} \and
George Retsinas\inst{2} \and
Vincent Christlein\inst{3} \and
Mathias Seuret\inst{3} \and
Giorgos Sfikas\inst{4,5} \and
Elisa Barney Smith\inst{1} \and
Hamam Mokayed\inst{1} \and
Marcus Liwicki\inst{1} }
\authorrunning{Nikolaidou et al.}
%
\institute{Luleå University of Technology, Sweden \\
\email{\{konstantina.nikolaidou, elisa.barney, hamam.mokayed, marcus.liwicki\}@ltu.se}\\
\and
National Technical University of Athens, Greece\\
\email{gretsinas@central.ntua.gr}
\and
Friedrich-Alexander-Universität, Germany\\
\email{\{vincent.christlein, mathias.seuret\}@fau.de}
\and
University of West Attica, Greece\\
\email{gsfikas@uniwa.gr}
\and
University of Ioannina, Greece\\
}
\maketitle              
%

\begin{abstract}

Text-to-Image synthesis is the task of generating an image according to a specific text description.
Generative Adversarial Networks have been considered the standard method for image synthesis virtually since their introduction.
Denoising Diffusion Probabilistic Models are recently setting a new baseline, with remarkable results in Text-to-Image synthesis, among other fields.
Aside its usefulness \emph{per se}, it can also be particularly relevant as a tool for data augmentation to aid training models for other document image processing tasks.
In this work, we present a latent diffusion-based method for styled text-to-text-content-image generation on word-level.
Our proposed method is able to generate realistic word image samples from different writer styles, by using class index styles and text content prompts without the need of adversarial training, writer recognition, or text recognition.
We gauge system performance with the Fr\'{e}chet Inception Distance, writer recognition accuracy, and writer retrieval.
We show that the proposed model produces samples that are aesthetically pleasing, help boosting text recognition performance, and get similar writer retrieval score as real data.
Code is available at: \url{https://github.com/koninik/WordStylist}.

%


\end{abstract}

\section{Introduction}
\label{sec:introduction}

Image synthesis is a very challenging problem in Computer Vision, which has gained traction with the rekindling of interest in neural networks a decade prior, and especially the introduction of models and concepts such as Generative Adversarial Networks (GANs)~\cite{NIPS2014_5ca3e9b1}, Variational Autoencoders (VAEs)~\cite{kingma2013auto} or Normalizing Flows (NFs) \cite{kingma2018glow}.
Apart from the utility of the generated image in itself, image synthesis has been employed as a tool to artificially augment training sets.
This is an aspect that is critical when it comes to training Deep Learning models, which are notorious for typically requiring vast amounts of data to attain optimal performance.
Annotating data is an expensive and time-consuming task that requires a lot of human effort and expertise.
A particular variant of image synthesis is text-to-image synthesis, where 
the task is to generate an image given a text description. 
As stated in~\cite{frolov2021adversarial}, a text description can indeed give more semantic and spatial information about the objects depicted in an image than a single label.
Text-to-image synthesis has been established as a whole independent field as several applications have gained relative prominence. 
\newline
\indent
Conditional Generative Adversarial Networks (cGANs) ~\cite{mirza2014conditional}, the conditional variant of GANs,  have further enabled the augmentation of existing datasets by generating data given a specific class or a specific input.
With the advent of these models, adversarial training has been established as the standard for image generation, where a minimax game is ``played'' between two networks, aptly named Generator and Discriminator.
The Generator is tasked with creating a sample -- in the current context, the synthesized image -- while the Discriminator is tasked with detecting instances that are outliers with respect to the training data.
Unlike GANs, which do not explicitly define a data density, other state-of-the-art approaches have attempted to approach data generation as sampling from a probability density function (pdf).
Variational Autoencoders cast the problem as one of estimating a latent representation for members of a given dataset, given the prior knowledge that latent embeddings are Gaussian-distributed.
They are comprised of two network parts, named the Encoder and the Decoder.
The Encoder produces (probabilistic) latent representations given a datum,
while the Decoder is tasked with the inverse task, that is producing a sample given a latent representation.
Normalizing flows also deal with estimating the pdf of a given set, and also assume the existence of a latent space that is to be estimated, like VAEs.
Latent data are equidimensional to the image data, and training is performed by learning a series of non-linear mappings that gradually convert the data distribution from and to a Gaussian distribution.
In VAEs as in NFs, once the model is trained, image generation can simply be performed by sampling from the latent space and applying the learned transformation back to the image / original space.
The outburst of Diffusion Models, and in particular more recent variants such as Denoising Diffusion Probabilistic Models (DDPMs) or Latent Diffusion Models (LDM) have quickly begun to change the picture of the state of the art
with achievements that can often be described to be no less than astonishing.
The results of systems
such as DALL$\cdot$E-2~\cite{Ramesh2022HierarchicalTI} and Imagen~\cite{Saharia2022PhotorealisticTD} have 
prompted many researchers to experiment with their use in different applications. 
Diffusion models \cite{SohlDickstein2015DeepUL} are based on a probabilistic framework like VAEs or NFs, but propose a different approach to the problem of image synthesis, cast in its standard form as density estimation followed by sampling.
Like NFs, in their standard form the latent space dimensionality is defined to be equal to that of the original space, and learning is performed by estimating a series of non-linear transformations between latent space and original space.
A ``forward/diffusion'' process gradually adds noise to inputs according to a predetermined schedule;
with the ``reverse'' process the aim is to produce an estimate of an image given a latent, noisy sample.

In this work, instead of using text only as a description of the image contents, 
we also use it literally as image content, in the sense of generating handwriting.
Thus, we address a task of Text-to-Text-Content-Image Synthesis.
The main contributions of this work are the following:
\begin{enumerate}
    \item We present a method based on a conditional Latent Diffusion Model, that takes as input a word string and a style class and generates a synthetic image containing that word.
    \item We compare qualitative results of our method with other GAN-based generative model approaches.   
    \item We further evaluate our results by presenting qualitative and quantitative results for text recognition using the synthetic data. The synthetic data is used for data augmentation, resulting in boosting the performance of a state-of-the-art Handwriting Text Recognition (HTR) system.
    \item And finally, we compare synthetic data and real handwritten paragraphs using a writer retrieval system. We show that data produced by our method show no significant difference in style to real data, and outperforms the other methods by a tremendous margin.
\end{enumerate}

The paper is organized as follows.
In Section \ref{sec:bg_related_work}, we present an overview of the related work. 
Our proposed method is introduced in Section \ref{sec:method}, while Section \ref{sec:experiments} includes the evaluation process and results.
Section \ref{sec:limitations} presents limitations and possible future directions.
Finally, we discuss conclusions in Section \ref{sec:conclusion}.

\section{Related Work}
\label{sec:bg_related_work}



\textbf{Text-to-Text-Content-Image synthesis} refers to the task of generating an image that depicts a specific text, whether it is on the character-, word-, sentence-, or page-level, given that text as the input condition.
A field directly related to this task is Document Image Analysis and Recognition, notably one of the resource-constrained domains with respect to the availability of annotated data, at least compared to the current state in natural image-related tasks~\cite{5206848,Lin2014MicrosoftCC}.

Most existing works focus on conditioning on a string prompt and a writer style to generate images of realistic handwritten text using GAN-based approaches.
GANwriting~\cite{kang2020ganwriting} creates realistic handwritten word images conditioned on text and writer style by guiding the generator.
The method is able to produce out-of-vocabulary words.
The authors extend this work in~\cite{kang2021content}, generating realistic handwritten text-lines.
SmartPatch~\cite{mattick2021smartpatch} fixes artifact issues that GANwriting faces by deploying a patch discriminator loss.
ScrabbleGAN~\cite{Fogel2020ScrabbleGANSV} uses a semi-supervised method to generate long handwritten sentences of different style and content.
A Transformer-based method is presented in~\cite{Bhunia_2021_ICCV}, using a typical Transformer Encoder-Decoder architecture that takes as inputs style features of handwritten sentence images extracted by a CNN encoder and a query text in the decoding part.
The model is trained with a four-part loss function, including an adversarial loss, a text recognition loss, a cycle loss, and a reconstruction loss.

Related to Historical Document Analysis~\cite{lombardi2020deep,Nikolaidou2022ASO}, the work presented in~\cite{8893115} initially generates modern documents using \LaTeX\ and then attempts to convert them into a historical style with the use of CycleGAN~\cite{zhu2017unpaired}.
The work is further extended in~\cite{vogtlin2021generating}, by adding text recognition to the framework and the loss function, which gives better readable text in the image synthesis.

\section{Method}
\label{sec:method}

In this section, we present some general background information for the standard Diffusion and Latent Diffusion Models. 
We then illustrate in detail the proposed method that includes the forward process, model components, sampling and experimental setup for training and sampling from the model.

\subsection{Diffusion Models Background}

\subsubsection{\acrfull{ddpm}.} 
Diffusion Models are a type of generative model that employ Markov chains to add noise and disrupt the structure of data. 
The models then learn to reverse this process and reconstruct the data. Inspired by Thermodynamics~\cite{SohlDickstein2015DeepUL}, Diffusion Models have gained popularity in the field of image synthesis due to their ability to generate high quality samples.

The Diffusion Model consists of two phases: the forward (diffusion) process and the reverse (denoising) process.
In the forward process, a sample $x_0$ is initially drawn from a 
distribution $x_0\sim q(x_0)$ corresponding to the observed data.
This is subjected to Gaussian noise, which produces a latent variable $x_1$;
noise is again added to $x_1$, giving latent variable $x_2$, and so on, until some predefined hyper-parameter $T$.
This process forms a series of latent variables $x_1, x_2, \cdot, x_T$,
Formally, we can write:
\begin{equation}
    q(x_{1:T}|x_0) = \prod_{t=1}^{T} q(x_t|x_{t-1}),\quad
    q(x_t|x_{t-1}) = N(x_t;\sqrt{1-\beta_t}x_{t-1}, \beta_tI),
\end{equation}
where we have $\beta_i \in [0, 1], \forall i \in [1,T]$.
Hyper-parameters $\beta_1,\beta_2,...,\beta_T$ collectively form 
a noise variance schedule, used to control the amount of noise added at each timestep.
In the final timestep, given large enough $T$ and suitable noise schedule, we will have $q(x_T|x_0) = q(x_T) \approx N(0, I)$, i.e. 
the end result becomes practically a pure Gaussian noise sample with no structure.
In the reverse (denoising) phase, a neural network learns to gradually remove the noise from the sampled by a stationary distribution until ending up with actual data.
Hence, image synthesis will be performed according to an ancestral sampling scheme.
This means that first we need to sample from $q(x_T)$, then we sample by the previous time-step conditioned on the sampled value of $x_T$, and so and so forth until we sample the required $x_0$.

The noise is gradually removed in reverse timesteps using the following transition:
\begin{equation}
\resizebox{.92\hsize}{!}{$
p_{\theta}(x_{0:T}) = p(x_T)\prod_{t=1}^{T} p_{\theta}(x_{t-1}|x_t),\quad p_{\theta}(x_{t-1}|x_t) = N(x_{t-1};\mu_\theta(x_t, t), \Sigma_\theta(x_t, t)) .$}
\end{equation}
The network is trained by optimizing the variational lower bound between the forward process posterior and the joint distribution of the reverse process $p_\theta$.
The training loss 
\begin{equation}
L = \mathbb{E}_{x_0,t,\epsilon}[||\epsilon - \epsilon_\theta(x_t, t)||^2]
\end{equation}
is calculated as the reconstruction error between the actual noise, $\epsilon$, and the estimated noise, $\epsilon_\theta$.
In the case of Latent Diffusion models the loss will be adapted to the latent representation $z_t$.

\subsubsection{Latent Diffusion Models (LDM).} Diffusion Models have demonstrated remarkable performance in image generation and transformation tasks~\cite{ho2020denoising,kingma2021variational,kong2020diffwave,mittal2021symbolic}.
However, their computational cost is high due to the size of the input data and the use of cross-attention in images.
To address this issue, Latent Diffusion Models were introduced in~\cite{rombach2022high} to model the data distribution in a lower-dimensional latent representation space.
This is accomplished by mapping the input images to a latent representation using an encoder, and then decoding the sampled latents back into an image using a decoder, both from a variational autoencoder architecture.

\subsection{Proposed Approach}

\begin{figure}[t] 
 \includegraphics[width=0.98\textwidth, scale=0.1]{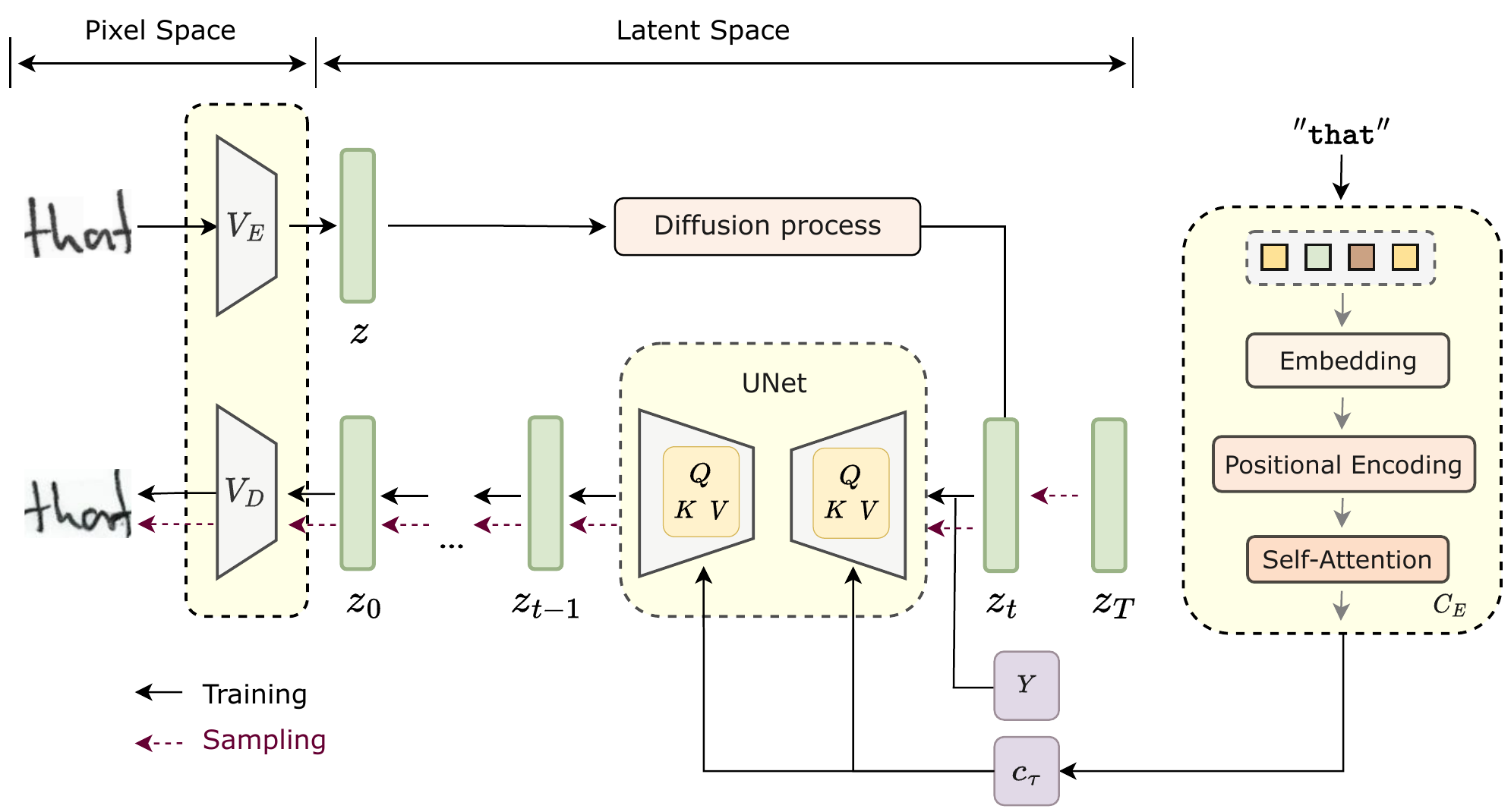}
    \centering
    \caption{\textbf{The overall architecture.} During training, an input image is fed to the encoder $V_E$ to create a latent representation $z$, then noise is added to the latent. The noisy latent $z_t$ is then fed to the U-Net noise predictor along with a style class index for the writer style and an encoded word as the content. The UNet predicts the noise of the noisy latent $z_{t-1}$ where $t-1$ is the corresponding timestep. During sampling, a random noise latent $z_T$ is given to predict its noise. Then the model uses the two noise predictions to reconstruct the latent of the image $z_0$ that is finally decoded by the decoder $V_D$ that creates the synthetic image.}
    \label{fig:model}
\end{figure}

The goal of this work is to generate synthetic word-image samples given a word string and a style class as conditions from a known distribution.
We approach this problem with the use of latent diffusion models to minimize training time and computational cost.
To move to the latent space we use the pre-trained ``stable-diffusion" VAE implementation from the Hugging Face repository\footnote{\url{https://huggingface.co/CompVis/stable-diffusion}}.
Figure \ref{fig:model} presents the overall architecture of the proposed method.

\subsubsection{Forward Process and Training.}
For the forward process, the VAE encoder $V_E$ initially transforms an input image to a latent representation $z$.
A diffusion model $p_\theta(x|Y,c_\tau)$ is learned on the style $Y$ and text-condition $c_\tau$ pairs.
Timesteps $t$ are sampled from a uniform distribution and the latent representation $z$ gets gradually corrupted by the diffusion process in every timestep.
For the noise prediction, we use a U-Net architecture~\cite{ronneberger2015u} with Residual Blocks~\cite{he2016deep} and intermediate Transformer Blocks~\cite{vaswani2017attention} to add the text condition to the model, as typically used by Ho et al.~\cite{ho2020denoising}.
The network takes as input the noisy image latents, the corresponding timestep, and the desired conditions $Y$ and $\tau$.
Timesteps are encoded using a sinusoidal position embedding, similar to~\cite{vaswani2017attention} to inform the model about each particular timestep that is operating.
The training objective is to minimize the reconstruction error between the network's noise prediction and the noise present in the image.
For the diffusion process, a noise scheduler increases the amount of noise linearly from $\beta_1=10^{-4}$ to $\beta_T=0.02$ for $T = 1000$ timesteps.
While most works use multiple ResNet blocks within the U-Net components, in the context of the current problem we need to take into account that we must work with scarce data compared to other use-cases;
larger models correspond to larger parameter spaces, which are exponentially harder to explore.
Hence, we use $1$ ResNet block in every module of the U-Net.
To further reduce the parameters and complexity of the network we use an inner model dimension of $320$ and $4$ heads in the Multi-Headed Attention layers within the U-Net.

\subsubsection{Sampling.}
\begin{figure}[ht]
\begin{center}

\centering
\renewcommand{\arraystretch}{1.5}
\addtolength{\tabcolsep}{2.5pt} 
\begin{tabular}{cccccccccc}


$T=100$ & $T=200$ &$T=300$ &$T=400$ &$T=500$ \\
\includegraphics[scale=0.25]{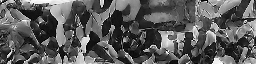} & \includegraphics[scale=0.25]{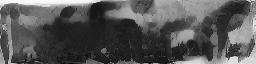} & \includegraphics[scale=0.25]{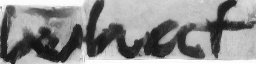} & \includegraphics[scale=0.25]{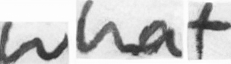} &
\includegraphics[scale=0.25]{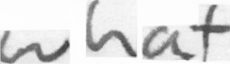}
\\
$T=600$ &$T=700$ &$T=800$ &$T=900$ &$T=1000$ \\
\includegraphics[scale=0.25]{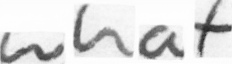}
&
\includegraphics[scale=0.25]{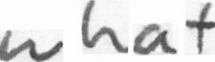}
&
\includegraphics[scale=0.25]{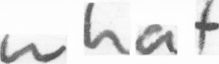}
&
\includegraphics[scale=0.25]{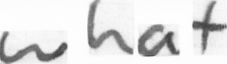}&
\includegraphics[scale=0.25]{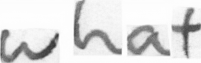}\\

\end{tabular}
\caption{Sampling outputs using various timesteps values in the reverse denoising process.}
\label{tab:timesteps}
\end{center}
\end{figure}

We generate synthetic samples by deploying the reverse denoising process learned from the model.
To this end, the noise of a random noisy sample $z_T$ is predicted by the learned network $p_\theta$ and gradually removed in every timestep of the reversed process starting from $T$ to $t=0$.
One of the main challenges associated with DDPM is the time required for sampling.
Our experiments indicate that reducing the number of time steps from 1,000 to 600 does not compromise the quality of the generated samples. 
The final image is obtained in pixel space by decoding the denoised latent variable using decoder $V_D$.
We demonstrate how the reduction of timesteps affects the quality of the generated sample in Figure \ref{tab:timesteps}.
The figure shows that below 500 timesteps the quality of the images is really affected, thus to make sure the generated samples are not affected dramatically we proceed with a value of 600.

\subsubsection{Style and Text Conditions.}
\begin{figure}[ht]
\begin{center}

\centering
\renewcommand{\arraystretch}{1.6}
\addtolength{\tabcolsep}{9.7pt} 
\begin{tabular}{lccc}

\hline
$Word$ $\tau$ & \texttt{"cool"} & \texttt{"About"} & \texttt{"kraut"}\\
\hline
- $C_E$ & \includegraphics[scale=0.3]{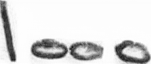} & \includegraphics[scale=0.3]{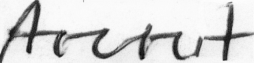} & \includegraphics[scale=0.4]{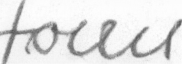} \\

+ $C_E$ &\includegraphics[scale=0.25]{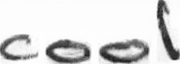} & \includegraphics[scale=0.35]{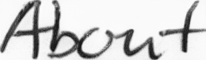} & \includegraphics[scale=0.45]{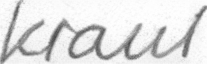} \\
\hline

\end{tabular}
\caption{Comparison of generated images with (top row) and without (bottom row) the positional encoding and self-attention layer of the context encoder $C_E$. All image pairs (top-bottom) share the same style condition.}
\label{fig:encoder}
\end{center}
\end{figure}

The input style condition $Y$ is processed with an embedding layer and then added to the timestep embedding.
For the text condition, a content encoder $C_E$ is used to transform an input string $\tau$ into a meaningful context representation $c_\tau = C_E(\tau)$ for the model.
Initially, the string is tokenized using a unique index for each letter and then passed through an embedding layer to transform it to an appropriate embedding dimension according to the vocabulary size which is the number of characters present in the training set.
Then, positional encodings similar to~\cite{vaswani2017attention} are used to inform the model about the character position in the sequence with the use of sine and cosine functions as $PE_{pos, i} = \sin(pos/1000^{i/emb\_dim})$ and $PE_{pos, i+1} = \sin(pos/1000^{i/emb\_dim})$, where $pos$ is the position of each letter in the sequence and $emb\_dim$ is $320$ as mentioned previously. 
Finally, to create the text input condition $c_\tau$ a dot-product attention layer is used, defined as $Att(Q,K,V) = softmax(\frac{QK^T}{\sqrt{d_k}})V$, to create a weighted sum of the character representations.
To support the choice of the positional text encoding we present a few samples as ablation with and without the positional encoding and self-attention layers in Figure \ref{fig:encoder}.

\subsection{Experimental Setup}

We conducted extensive experiments using the IAM offline handwriting database on word-level~\cite{marti2002iam}.
Similar to \cite{mattick2021smartpatch} and \cite{kang2020ganwriting}, we used the Aachen split train set and included words of 2-7 characters to train the diffusion model.
Thus, during training the model sees 339 writer styles and approximately 45K words.
For consistency, all images were resized to a fixed height of $64$ pixels, retaining their aspect ratio. To handle variations in width, images of width smaller than $256$ pixels were center-padded, while larger ones were resized to the maximum width. Since the maximum number of characters is $7$, this resizing did not cause significant distortions in the images. Moreover, these images were intended for training other models, which could eventually lead to resizing or modifications of the original images.
AdamW~\cite{loshchilov2017decoupled} is used as the optimizer during training with a learning rate of $10^{-4}$.
To better understand the nature of the model, no augmentation is used on the images during training.
Each model was trained for 1K epochs with a batch size of 224 on a single A100 SXM GPU.


\section{Evaluation and Results}
\label{sec:experiments}

We evaluate the quality of the generated word-images using our method in three aspects: visual quality, text quality, and style quality. 
To assess visual quality, we compute the commonly used Fréchet Inception Distance (FID) score and provide examples of in-vocabulary and out-of-vocabulary words. 
Additionally, we demonstrate the results of blending two distinct writer styles through interpolation.
To determine the effectiveness and text quality of our approach, we create a pseudo training set from the IAM database and conduct several experiments for handwriting text recognition (HTR). 
For comparison with other methods, we perform the same experiments using two GAN-based approaches, SmartPatch and GANwriting.
Finally, we evaluate style quality in two ways.
First, we train a standard Convolutional Neural Network (CNN) on the real IAM database for style classification and test it on the generated samples.
Second, we apply a writer retrieval method and compare its performances using real or synthetic data.
This enables us to measure the extent to which our method accurately captures the style of the original IAM database.

\subsection{Qualitative Results}

\begin{figure}[ht]
\begin{center}

\centering
\renewcommand{\arraystretch}{1.6}
\addtolength{\tabcolsep}{2.7pt} 
\begin{adjustbox}{scale=1.,center}
\begin{tabular}{cccc}
\hline
Real IAM & WordStylist (ours) & SmartPatch & GANwriting\\
\hline

\includegraphics[scale=0.3]{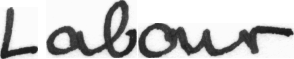} & \includegraphics[scale=0.3]{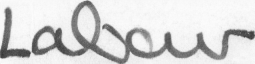} & \includegraphics[scale=0.4]{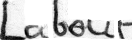} & \includegraphics[scale=0.4]{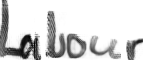}\\

\includegraphics[scale=0.25]{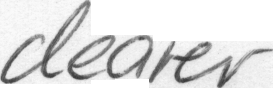} & \includegraphics[scale=0.35]{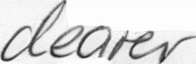} & \includegraphics[scale=0.45]{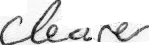} & \includegraphics[scale=0.4]{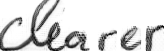}\\

\includegraphics[scale=0.4]{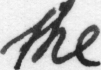} & \includegraphics[scale=0.4]{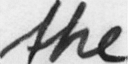} & \includegraphics[scale=0.4]{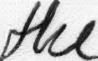} & \includegraphics[scale=0.4]{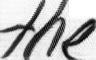}\\

\includegraphics[scale=0.27]{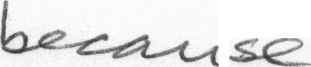} & \includegraphics[scale=0.3]{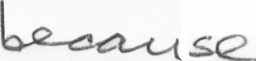} & \includegraphics[scale=0.4]{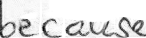} & \includegraphics[scale=0.4]{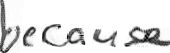}\\

\includegraphics[scale=0.3]{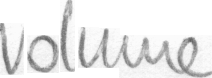} & \includegraphics[scale=0.35]{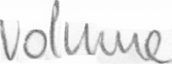} & \includegraphics[scale=0.47]{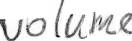} & \includegraphics[scale=0.42]{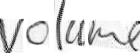}\\

\hline
\end{tabular}
\end{adjustbox}
\caption{Comparison of (in-vocabulary) real word images and synthetic versions of these words.}
\label{tab:comparison_gen}
\end{center}
\end{figure}

A comparative qualitative evaluation can be found in Figure \ref{tab:comparison_gen}, where both SmartPatch and GANwriting methods have been used to generate a set of word images. 
Specifically, the goal was to recreate the original images (further left column).
As we can see, all methods generate ``readable" words without notable artifacts/deformations. 
Nonetheless, SmartPatch has a smoother appearance compared to GANwriting, as it was designed to do, while the proposed Diffusion approach retains the original style to an outstanding degree. 

\begin{figure}[ht]
\begin{center}

\centering
\renewcommand{\arraystretch}{1.3}
\addtolength{\tabcolsep}{.7pt} 
\begin{adjustbox}{scale=0.75,center}
\begin{tabular}{ccccccccc}
\toprule

 \multicolumn{4}{c}{In-Vocabulary (IV) Generated Words} & &\multicolumn{4}{c}{Out-of-Vocabulary (OOV) Generated Words} \\
 \cmidrule(r){1-4} \cmidrule(r){6-9}
 \includegraphics[scale=0.25]{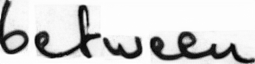} & \includegraphics[scale=0.25]{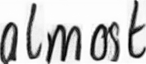} & \includegraphics[scale=0.25]{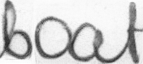}& \includegraphics[scale=0.25]{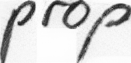}  & 
 
 & \includegraphics[scale=0.23]{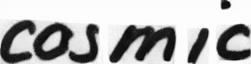}& \includegraphics[scale=0.25]{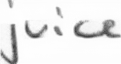} & \includegraphics[scale=0.25]{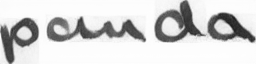} 
 & \includegraphics[scale=0.26]{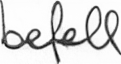} 
\\

 \includegraphics[scale=0.29]{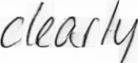}& \includegraphics[scale=0.18]{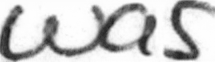} & \includegraphics[scale=0.25]{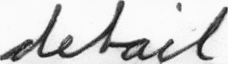} 
 & \includegraphics[scale=0.35]{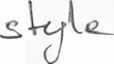}  & &\includegraphics[scale=0.25]{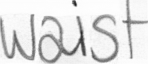}& \includegraphics[scale=0.26]{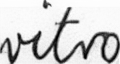} & \includegraphics[scale=0.25]{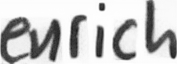}& \includegraphics[scale=0.3]{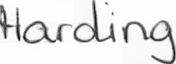}  
\\
 \includegraphics[scale=0.25]{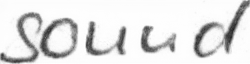}& \includegraphics[scale=0.33]{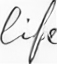}  & \includegraphics[scale=0.25]{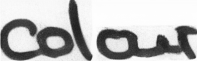} & \includegraphics[scale=0.25]{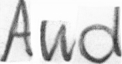} & & 
 
 \includegraphics[scale=0.25]{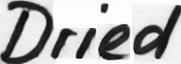}& \includegraphics[scale=0.25]{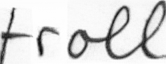} & \includegraphics[scale=0.25]{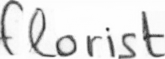}& \includegraphics[scale=0.22]{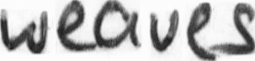}  
\\

 \includegraphics[scale=0.35]{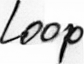}& \includegraphics[scale=0.33]{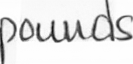}  & \includegraphics[scale=0.3]{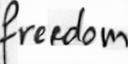} & \includegraphics[scale=0.25]{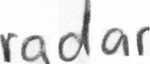} & & 
 
 \includegraphics[scale=0.25]{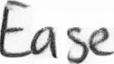}& \includegraphics[scale=0.25]{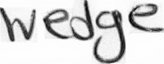} & \includegraphics[scale=0.25]{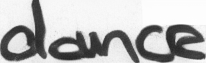}& \includegraphics[scale=0.3]{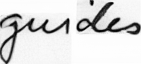}  
\\
\bottomrule
\end{tabular}
\end{adjustbox}
\caption{Qualitative results from WordStylist of random writer styles from In-Vocabulary (IV) (left) and Out-of-Vocabulary (OOV) (right) word generation. All writer styles are randomly selected to produce each word meaning that the IV samples may not appear in the training set with the presented style-text combination.}
\label{fig:iv_oov}
\end{center}
\end{figure}

Furthermore, to validate the variety of styles and the ability to generalize beyond already seen words, in Figure \ref{fig:iv_oov} we present generated samples using our method of In-Vocabulary (IV) and Out-of-Vocabulary (OOV) words and random styles picked from the IAM training set.
We can observe a notable variety over the writing style, indicating the good behavior of the proposed method, even for the case of OOV words that were never met in the training phase.


Towards measuring the quality of the generation,
a metric commonly used to evaluate generative models is the Fréchet Inception Distance (FID) score~\cite{dowson1982frechet}.
The metric computes the distance between two dataset feature vectors extracted by an InceptionV3 network~\cite{szegedy2016rethinking} pre-trained on ImageNet~\cite{5206848}.
Our approach achieves an \textbf{FID} score of \textbf{22.74}, which is comparable to SmartPatch's score of \textbf{22.55}. 
GANwriting performs with an FID of \textbf{29.94}. While FID is a widely used metric for evaluating generative models, it may not be appropriate for tasks that do not involve natural images similar to those in ImageNet, on which the network was trained.
In fact, this domain shift between natural images and handwritten documents lessens the fidelity of the evaluation protocol, but adapting this metric, by fine-tuning the FID network on document images, is out of the scope of this work. Despite this, the FID metric is still an indication of realistic images.


\subsection{Latent Space Interpolation}
\begin{figure}[ht]
\begin{center}

\centering
\renewcommand{\arraystretch}{1.8}
\addtolength{\tabcolsep}{0.5pt} 
\begin{adjustbox}{scale=0.72,center}
\begin{tabular}{cccccccc}
\hline
$Y_A$ & 0.0 & 0.2 & 0.4 &  0.6 & 0.8 & 1.0 & $Y_B$\\
\hline

 Real A &  &  &  &  &  &  & Real B\\
 
 \includegraphics[scale=0.32]{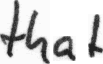} & \includegraphics[scale=0.3]{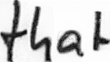}  & \includegraphics[scale=0.3]{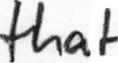} & \includegraphics[scale=0.3]{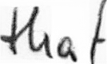}  & \includegraphics[scale=0.3]{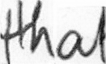} & \includegraphics[scale=0.3]{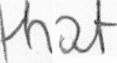} & \includegraphics[scale=0.3]{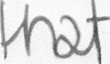} &
\includegraphics[scale=0.2]{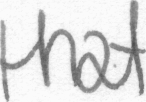}\\
\includegraphics[scale=0.2]{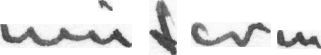} & \includegraphics[scale=0.2]{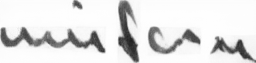}  & \includegraphics[scale=0.2]{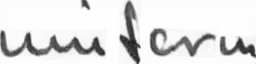} & \includegraphics[scale=0.2]{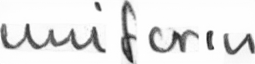}  & \includegraphics[scale=0.2]{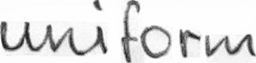}  & \includegraphics[scale=0.2]{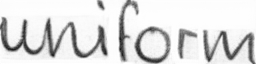}  & \includegraphics[scale=0.2]{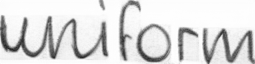} &
\includegraphics[scale=0.2]{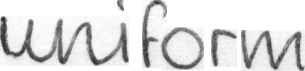}\\

\includegraphics[scale=0.2]{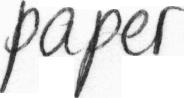} & \includegraphics[scale=0.3]{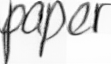}  & \includegraphics[scale=0.3]{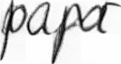} & \includegraphics[scale=0.3]{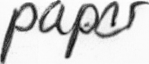}  & \includegraphics[scale=0.3]{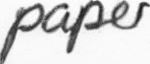}  & \includegraphics[scale=0.3]{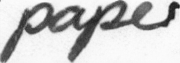}  & \includegraphics[scale=0.3]{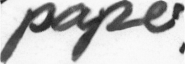} &
\includegraphics[scale=0.3]{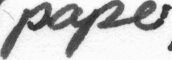}\\

 \includegraphics[scale=0.2]{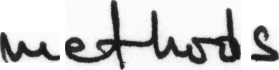} & \includegraphics[scale=0.2]{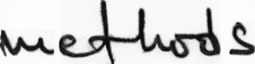}  & \includegraphics[scale=0.2]{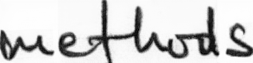} & \includegraphics[scale=0.2]{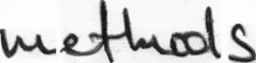} & \includegraphics[scale=0.2]{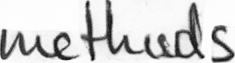} & \includegraphics[scale=0.2]{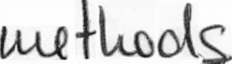}  & \includegraphics[scale=0.2]{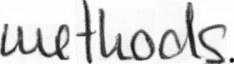} &
\includegraphics[scale=0.2]{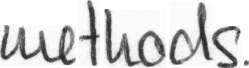}\\

 \includegraphics[scale=0.2]{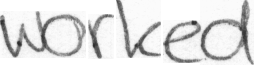} & \includegraphics[scale=0.2]{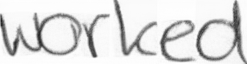}  & \includegraphics[scale=0.2]{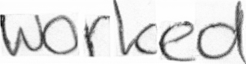} & \includegraphics[scale=0.3]{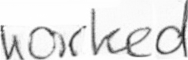}  & \includegraphics[scale=0.3]{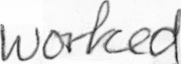}  & \includegraphics[scale=0.3]{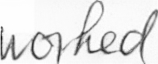}  & \includegraphics[scale=0.3]{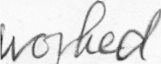} &
\includegraphics[scale=0.2]{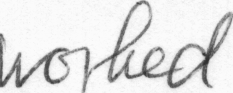}\\

\hline
\end{tabular}
\end{adjustbox}
\caption{Interpolation results between writer styles with various weights.}
\label{tab:interpol}
\end{center}
\end{figure}

Following the paradigm of GANwriting~\cite{kang2020ganwriting}, we further interpolate between two writer styles $Y_A$ and $Y_B$ by a weight $\lambda_{AB}$ to create mixed styles.
Using a weighted average $Y_{AB} = (1-\lambda_{AB})Y_A + \lambda_{AB}Y_B$, we interpolate between $Y_A$ and $Y_B$ for a fixed text condition.
Figure~\ref{tab:interpol} shows the results on fixed words with interpolation between two writing styles with various $\lambda_{AB}$ values.
One can observe the smooth progression between styles as the mix parameter $\lambda_{AB}$ increases.
This interpolation concept could be a useful tool for generating words of unseen/unknown style, especially if the goal is to create an augmented dataset for training document analysis methods.

\subsection{Handwriting Text Recognition (HTR)}
We evaluate the generated data on the task of Handwriting Text Recognition and assess the usefulness of the data on a standard downstream task.
We use the HTR system presented in~\cite{10.1007/978-3-031-06555-2_17}.
Specifically, the used HTR system is a hybrid CNN-LSTM network with a ResNet-like CNN backbone followed by a 3-layers bi-directional LSTM head, trained with Connectionist Temporal Classification (CTC) loss~\cite{graves2008novel}. 
We followed the modifications proposed in \cite{10.1007/978-3-031-06555-2_17} and used a column-wise max-pooling operation between the CNN backbone and the recurrent head, as well as a CTC shortcut of a shallow 1D CNN head. 
This shortcut module, as described in the initial work, is discarded during testing and is used only for assisting the training procedure.
Input word images have a fixed size of $64 \times 256$ by performing a padding operation (or resized if they exceed the pre-defined size).

For comparison with the related work on word-image generation, we further evaluate GANwriting and SmartPatch on the HTR task with the same model.

Using the generated images to train an HTR system and then evaluate the trained system on the original test set of real images aims to a multifaceted insight on the quality of the data; Achieving good results in the test set translates to ``readable" words (at least in their majority), so that the system can understand the existing characters during training with CTC, as well as to a variability in writing styles, so that the training system could generalize well in the test set of unseen writing styles. The ideal generative model should abide to both these properties and thus can be used to train a well-performing HTR system.

Following the protocol of~\cite{mattick2021smartpatch}, for this recognition task, we discarded, both from the training and the test set, words containing non-alphanumeric characters, as well as words with more than 10 characters, since the generative models have been trained considering the same setup.
We used the generative models to recreate the train set, both in text and in style. 
The results of this experiment are reported in Table~\ref{HTR_results}, where we present the character error and word error rates (CER/WER) for the initial IAM train set, the recreated sets of the generative models (i.e., GANwriting, SmartPatch and our proposed Diffusion approach), as well as the combination of the original set with each one of the recreated (i.e., with $\times2$ training images, compared to the initial set).
The reported results correspond to the mean value and the standard deviation over 3 different training/evaluation runs for each setup.
The following observations can be made:
\begin{itemize}
    \item The generated synthetic datasets under-perform with respect to the original IAM dataset. However, both GANwriting and SmartPatch approaches lead to a notable decrease in performance, indicating lack of writing style variability. 
    On the other hand, the proposed method achieves considerably low error rates, but not on par with the real data.
    \item Combining the synthetic datasets with the real IAM train set, the performance is improved compared to training only on the original IAM set, with the exception of GANwriting and the CER metric, which is practically on par with the baseline model.
    \item SmartPatch, despite visually improving the results of GANwriting, does little to improve the HTR performance.
    \item The synthetic set, generated by our proposed method, along with the real set, considerably outperforms all other settings and is statistically significant with a $p$ value of 0.035. 
\end{itemize}

\begin{table}
\caption{HTR results, reporting the Character Error Rate (CER) and Word Error Rate (WER). For both metrics, the lower the better.}\label{HTR_results}
\begin{adjustbox}{scale=1.15,center}
\centering
\renewcommand{\arraystretch}{1.3}
\addtolength{\tabcolsep}{2.5pt}  
\begin{tabularx}{0.82\textwidth}{lcc}

\hline
\textbf{Training Data} &  \textbf{CER (\%) $\downarrow$} & \textbf{WER (\%) $\downarrow$} \\
\hline

Real IAM & \phantom{3}$4.86\pm0.07$& $14.11\pm0.12$ \\

GANwriting IAM  &  $38.74\pm0.57$&  $68.47\pm0.32$ \\

SmartPatch IAM  &  $36.63\pm0.71$ & $65.25\pm1.02$ \\

WordStylist IAM (Ours) & \phantom{3}$8.80\pm0.12$  & $21.93\pm0.17$\\

Real IAM + GANwriting IAM   & \phantom{3}$4.87\pm0.09$ & $13.88\pm0.10$ \\

Real IAM + SmartPatch IAM  & \phantom{3}$4.83\pm0.08$ & $13.90\pm0.22$ \\

Real IAM + WordStylist IAM (Ours)  & \phantom{3}$\textbf{4.67}\pm\textbf{0.08}$ & $\textbf{13.28}\pm\textbf{0.20}$ \\

\hline
\end{tabularx}
\end{adjustbox}
\end{table}

\subsection{Handwriting Style Evaluation}
\label{ssct:style-evaluation}

Qualitative results show that our proposed method is able to nicely capture the style of each writer present in the IAM database.
In order to quantify this property, we employ an implicit evaluation via writer identification.

\begin{figure}[tb]
\centering
\subfloat[Real IAM.]{\label{4figs-a} \includegraphics[width=0.4\textwidth]{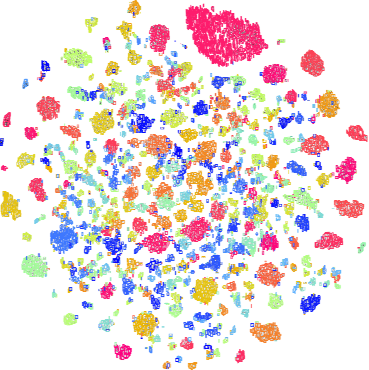}}
\hfill
\subfloat[WordStylist IAM (ours).]{\label{4figs-b} \includegraphics[width=0.4\textwidth]{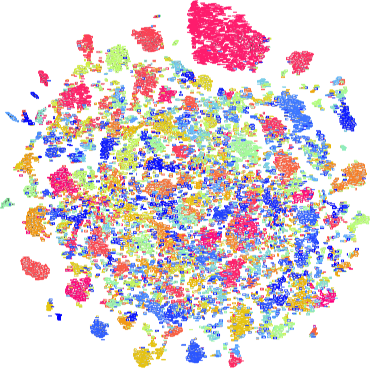}}%
\hfill
\subfloat[SmartPatch IAM.]{\label{4figs-c} \includegraphics[width=0.4\textwidth]{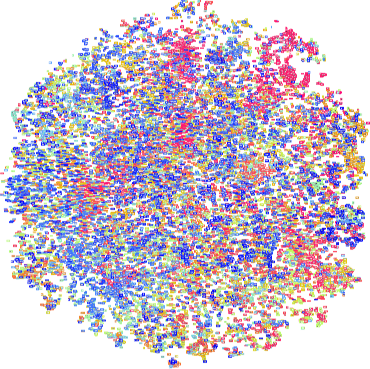}}%
\hfill
\subfloat[GANwriting IAM.]{\label{4figs-d} \includegraphics[width=0.4\textwidth]{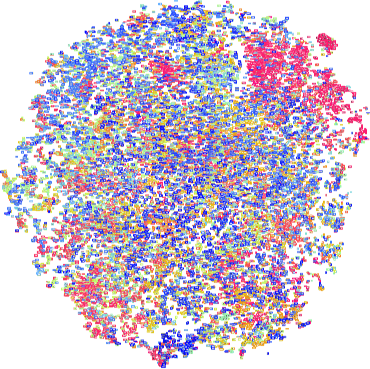}}%
\caption{T-SNE projections of the feature vector produced by the ResNet18 trained for writer identification, as described in Section~\ref{ssct:style-evaluation}.}
\label{tsne}
\end{figure}

The most straightforward way to address this is via a writer classification formulation. Specifically, to evaluate the generated styles, we finetuned a ResNet18 CNN~\cite{he2016deep}, pre-trained on ImageNet, on the IAM database for the task of writer classification.
Then, we use the generated datasets from the three generative methods as test sets and present the obtained accuracy in Table \ref{style_accuracy}.
The network manages to successfully classify most of the generated samples from our proposed method with an accuracy of 70.67\%, while it fails to recognize classes on samples from the other two methods.
This result comes as no surprise since the proposed method learns explicitly the existing styles, while both the GAN-based approaches adapt the style based on a few-shot scheme. 
Furthermore, we use the features extracted by the model to plot t-SNE embeddings on the different datasets in Figure \ref{tsne}.
In more detail, we used the 512-dimensional feature vector extracted by the second-to-last layer, trying to simulate a style-based representation space. 
Again, the resulted projection of the data generated by our Diffusion approach appears to be much closer to the real data.
On the contrary, the GAN-based methods create ``noisy`` visualization with no distinct style neighborhoods.
In fact, even the proposed method seems to have a similar noisy behavior (in the center of the plot) but to a much lesser extent. 
This phenomenon is in line with the HTR results, where the diffusion method provided results much closer to the real IAM, but not on par.
\newline
\indent
As an alternative to the straightforward implementation of writer classification, we also use a classic writer retrieval pipeline consisting of local feature extraction and computing a global feature representation~\cite{Christlein15ICDAR,Christlein17PR,Christlein18DAS}.
While the local descriptors can also be trained in a self-supervised~\cite{Christlein17ICDAR}, we just use SIFT~\cite{Lowe04} descriptors extracted on SIFT keypoints. 
The descriptors are normalized using Hellinger normalization~\cite{Arandjelovic12} (a.\,k.\,a.\ as RootSIFT) and are subsequently jointly whitened and dimensionality-reduced using PCA~\cite{Christlein17PR}. 
The global feature representation is computed using multi-VLAD~\cite{Christlein15ICDAR}, where the individual VLAD representations use generalized max-pooling~\cite{Christlein18DAS}.
\newline
\indent
This pipeline needs paragraphs as input in order to gather a sufficient amount of information.
To produce synthetic text paragraphs, we paste randomly-selected synthetic words on a blank background, following a similar structure as the printed text of IAM: same number of lines, similar number of characters per line.
Thus, no information from the handwritten text is used.
Line spacing is constant, and a small randomness is added to word spacing.
\newline
\indent
We use a leave-one-image-out cross-validation, i.\,e., each sample is used as query and the results are averaged. 
As metrics, we give the top-1 accuracy and mean average precision (mAP).
For our experiment, we use two paragraphs of 157 writers (IAM + IAM). 
In subsequent experiments, we replace the second paragraph by the synthesizers (GANWriting, SmartPatch, WordStylist). 
In this way, the query sample is either an original sample and the closest match should be the synthetic one or vice-versa.
\newline
\indent
The results, given in Table~\ref{tab:writer_retrieval}, show little difference between real data (IAM + IAM) and data produced by our method (IAM + WordStylist).
Thus, our method produces persistent writing styles that are nearly indistinguishable for the writer retrieval pipeline.
It is able to imitate handwriting much better than GANwriting and SmartPatch, which both achieve significantly lower scores in this experiment.

\begin{table}[t]
\caption{Classification accuracy of a ResNet18 trained for writer identification on real data.}\label{style_accuracy}
\centering
\renewcommand{\arraystretch}{1.3}
\addtolength{\tabcolsep}{1.5pt}  
\begin{tabularx}{0.46\textwidth}{lc}
\toprule
\textbf{Test Set} & \textbf{Accuracy (\%)\textuparrow}  \\
\midrule
GANwriting &  \phantom{7}4.81\\
SmartPatch & \phantom{7}4.09 \\
WordStylist (Ours) & \textbf{70.67} \\
\bottomrule
\end{tabularx}
\end{table}

\begin{table}[t]
    \centering
    \caption{Writer retrieval results using a 157 writers subset of IAM.}
    \label{tab:writer_retrieval}
    \begin{tabular}{lc@{\hspace{0.005cm}}c}
    \toprule
                    & \textbf{Top-1 [\%] \textuparrow} & \textbf{mAP [\%] \textuparrow}\\
                                \midrule
         IAM + IAM              & 97.45 & 97.61\\
         \midrule
         IAM + GANwriting       & \phantom{9}3.18  & \phantom{9}7.23\\
         IAM + SmartPatch       & \phantom{9}3.18  & \phantom{9}7.72\\
         IAM + WordStylist (Ours) & \textbf{97.13} & \textbf{97.84}\\
         \bottomrule
    \end{tabular}
\end{table}






\section{Limitations and Future Work}
\label{sec:limitations}

Here, we address the limitations of the proposed method that pave the way towards future directions. 
We identified two main limitations in the proposed method:
\begin{itemize}
    \item \emph{Style Adaptation:} 
    Our proposed method, contrary to the compared GAN-based methods~\cite{kang2020ganwriting,mattick2021smartpatch}, explicitly takes the writing style as an input embedding. 
    This way, the model can learn to recreate such styles accurately.
    Nonetheless, adaptation to new styles is not straightforward with this pipeline. 
    The interpolation concept is a work-around to generate ``new'' styles, and can be extended to even interpolating K different styles. 
    Nonetheless, even such ideas do not provide the ability to adapt to a specific given style via few word examples, as done in~\cite{kang2020ganwriting}. 
    An interesting future direction is the projection of different style embeddings to a common style representation space, using deep features extracted by a writer classification model as done in Section~\ref{sec:experiments}.  
    \item \emph{Sampling Complexity:} Generating realistic examples requires many iterations (timesteps) in the sampling process. To generate a single image requires $\sim12~$sec, when using $T=600$, making the creation of large-scale datasets impractical.
    We aim to explore ways to assist the generation of quality images in fewer steps, while also utilizing a more lightweight network to further reduce the time requirements of a single step.
    \item \emph{Fixed Image Size:}
    As the generation process is initiated by sampling a latent Gaussian noise of a fixed shape, our proposed method currently generates images of a fixed shape. Generating text images of arbitrary shapes is a possible future direction to explore.
\end{itemize}

\section{Conclusion}
\label{sec:conclusion}

We presented WordStylist, a latent diffusion-based system for styled text-to-text-content image generation on word-level.
Our model manages to capture the style and content without the use of any adversarial training, text recognition and writer identification.
Qualitative and quantitative evaluation results show that our method produces high quality images which outperform significantly the state-of-art systems used for comparison.
Also we show that our synthetic word images can be used as extra training data to improve HTR accuracy.
The verisimilitude of the synthetic handwriting styles is proven by two experiments.
Using a CNN for writer identification, we obtain a classification accuracy of 70\% with our synthetic data, while the other generative methods used for comparison do not get higher than 5\%.
Also, t-SNE projections of the features learned by the CNN exhibit structures very similar to real data in the case of the proposed method only.
Moreover, we showed that using a recognized writer retrieval pipeline, there is no significant difference between results on our synthetic data and real data, both having a mAP slightly below 98\%.
The other generative methods do not perform as well, obtaining mAP below 8\%.
For future work, we aim to investigate the parameters and sampling of the model.
We further plan on extending this work for sentences and whole pages, focusing also on the layout of the document.

\section*{Acknowledgments}
We would like to thank Onkar Susladkar and Gayatri Deshmukh for their preliminary involvement in the early stages of the paper. 
\bibliographystyle{splncs04}
\bibliography{bibliography}

\end{document}